# EMS-NET: EFFICIENT MULTI-TEMPORAL SELF-ATTENTION FOR HYPERSPECTRAL CHANGE DETECTION

*Meiqi Hu[1], Chen Wu*[1], Bo Du[2]*

[1]State Key Laboratory of Information Engineering in Surveying, Mapping, and Remote Sensing, Wuhan University, Wuhan 430079, P. R. China
[2]School of Computer, Wuhan University, Wuhan 430072, P. R. China

## ABSTRACT

Hyperspectral change detection plays an essential role of monitoring the dynamic urban development and detecting precise fine object evolution and alteration. In this paper, we have proposed an original Efficient Multi-temporal Self-attention Network (EMS-Net) for hyperspectral change detection. The designed EMS module cuts redundancy of those similar and containing-no-changes feature maps, computing efficient multi-temporal change information for precise binary change map. Besides, to explore the clustering characteristics of the change detection, a novel supervised contrastive loss is provided to enhance the compactness of the unchanged. Experiments implemented on two hyperspectral change detection datasets manifests the out-standing performance and validity of proposed method.

## 1. INTRODUCTION

Change detection (CD) has been a classical challenging task for decades, aiming at detecting the difference from the multi-temporal remote sensing images acquired at the same location of different time [1]. Hyperspectral change detection provides unprecedent potential to discriminate the subtle and detailed changes [2], [3] for hyperspectral imaging sensors cover a larger frequency spectrum in narrow, continuous bands [4]. Nowadays, CD has been widely used in land cover and land usage monitoring, road and building damage detection under emergency rescue, national defense security monitoring [5], [6].

Currently, the deep learning-based methods have turn to the mainstream change detection approaches. Numerous siamese deep neural networks come forward for hyperspectral change detection [7]. [8] proposed an end-to-end Siamese CNN with spectral–spatial-wise attention to emphasize informative channels and locations. [9] put forward a multipath convolutional long short-term memory neural network learn multiscale temporal–spatial–spectral features, where a siamese CNN was adopted at first to extract spatial–spectral features. [10] developed a deep multiscale pyramid network to mine multilevel and multiscale spatial–spectral features, which were aggregated the level by level.

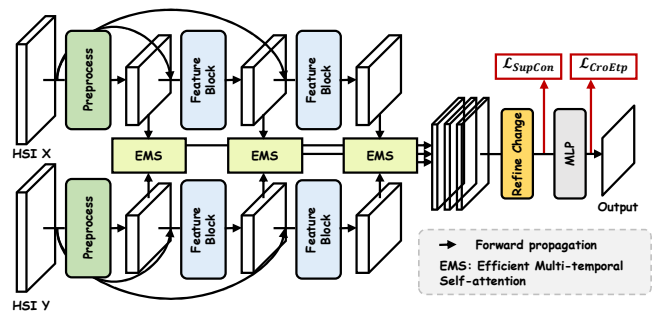

**Fig. 1** The proposed EMS-Net for hyperspectral change detection.

[11] designed a hyperspectral spatial-spectral feature understanding network, where a powerful spatial-spectral attention module is put forward to explore the spatial correlation and discriminative spectral features of multi-temporal hyperspectral images (HSI). The presented methods mentioned above provided abundant advanced spatial and spectral feature mining techniques while the exploration of the temporal change information is limited.

With the successive success of transformer in the fields of natural language processing and natural image [12], [13], the strong ability of sequence information excavation of transformer has attracted considerable researchers of remote sensing image processing. Transformer is capable of modeling powerful sequence relationship by the multi-head self-attention mechanism and promote the feature extraction ability by the multiple linear perceptron layers. And the advantage of global receptive field boosts the long-range relationship dependency. [14] put forward a joint spectral, spatial, and temporal transformer for hyperspectral image change detection, where the siamese spectral-spatial transformer firstly encoded the positional and spectral sequences and the temporal transformer is used to extract useful CD features. [15] presented a novel cross-temporal interaction symmetric attention network to integrate the difference features oriented from each temporal feature embedding. However, the massive computation of the self-attention brings a huge burden to the amount of network parameters and computation of these methods.

In brief, the current algorithms suffer from the inadequate exploitation of multi-temporal change information and the computation burden. To deal with these problems, an effective multi-temporal change information extraction approach is extremely urgent. Inspired by this, we proposed an innovative Efficient Multi-temporal Self-attention Network (EMS-Net) to acquire accurate hyperspectral change detection map. Several convolutional feature extraction blocks are firstly designed to extract multi-scaled discriminative features from bi-temporal HSIs. And EMS module is especially developed to build the correlation between different feature maps and fuse the closely related feature maps to reduce the redundancy. The multi-level change information is then concatenated to fed into binary change classifier to get the final change map. In addition, a hybrid loss function is presented to promote the accuracy of the unchanged class and the changed class. The EMS-Net aims at extracting efficient multi-temporal change features to detect precise change detection result under limited samples.

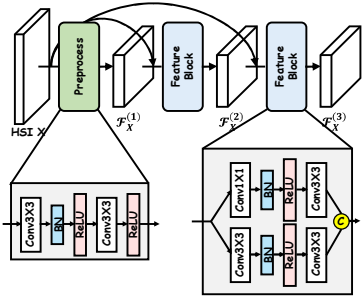

**Fig. 2.** The details of the Preprocess and Feature Block.

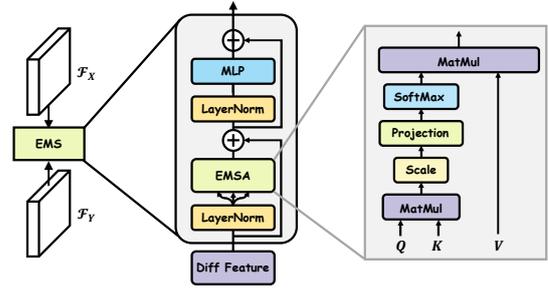

**Fig. 3.** The architecture of designed EMS module.

The paper is organized as follows. Section II will present the proposed method EMS-NET. The experiment results and analysis will be exhibited on Section III. And Section IV will conclude this paper.

## 2. METHODOLOGY

### 2.1 Framework

As Fig. 1 show, the proposed EMS-Net is a deep Siamese network for hyperspectral change detection. The HSIs are firstly processed by the pre-process module and another two feature blocks to extract multi-scaled spatial spectral features. And these abundant features full of geometric and sematic information are beneficial for dense prediction task like change detection. The designed Efficient Multi-temporal Self-attention module is tailored to integrate the extracted bi-temporal scaled features and excavate change information. The Refine Change module further explore the deep change features, which is finally inputted into the Multiple Linear Perceptron (MLP) served as binary change classifier. The loss function consists of two parts, the supervised contrastive loss $\mathcal{L}_{SupCon}$ and cross entropy loss $\mathcal{L}_{CroEtp}$. For better compactness of the unchanged class, we have designed an original supervised contrastive loss to constrain the feature embedding of the unchanged ones in the feature space.

### 2.2 EMS-NET

The two Siamese branches share the same weight for the convenience of comparing the similarity of bi-temporal HSIs. The details of backbone are presented as Fig. 2. Various features are extracted by different scaled convolutions. In addition, residual connections bridge the omissions of information from shallow layers to deep layers, adding sufficient details. The feature maps acquired from different layers correspond to diverse reception field, where the shallow features provide precise edge information and deep features offer high-level sematic information. All of these scaled features are vital for accurate dense prediction result.

Noted that there are a great number of similar feature maps between the bi-temporal HSIs features for plenty of unchanged ground objects. Therefore, the current concatenation of bi-temporal feature maps leads to tremendous redundant feature maps of the same ground objects. Additionally, the difference of the bi-temporal features yields redundant feature maps containing no change information. To deal with the current problems, the EMS module is proposed aiming at extracting multi-temporal change detection information efficiently from considerable high-dimensional feature maps of the two branches. Fig. 3 depict the detailed architecture of EMS module.

Mathematically, given feature maps extracted from bi-temporal HSIs as $\mathcal{F}_X$ and $\mathcal{F}_Y$, the difference feature map $\mathcal{F}_d$ is obtained by equation (1). To build the correlation between diverse difference feature maps, a novel EMSA is tailor to explore the redundancy of high-dimensional feature maps and fuse the closely related feature maps. The standard self-attention mechanism can be denoted as equation (2). However, for the input difference feature, the standard attention cannot discriminate the useless feature maps and redundant feature maps containing no change information still exist. So how to reduce the redundant features adaptively and obtain effective feature maps? As equation (3) show, the $\mathcal{W} \in \mathbb{R}^{C' \times C}$ is a distill weight matrix, where $C' < C$, $C$ is the number of spectral bands, targeted at distilling the abundant change information and cutting the useless and redundant feature maps. Consequently, the EMS module outputs distilled multi-temporal change feature maps, preserving the valuable change information and cutting the redundancy.

$$\mathcal{F}_d = \text{Abs}(\mathcal{F}_X - \mathcal{F}_Y) \quad (1)$$

$$Attention = \text{softmax}(\frac{Q \cdot K^T}{\sqrt{d}}) \cdot V \quad (2)$$

$$EMSA(Q, K, V) = \text{softmax}(\mathcal{W} \cdot \frac{Q \cdot K^T}{\sqrt{d}}) \cdot V \quad (3)$$

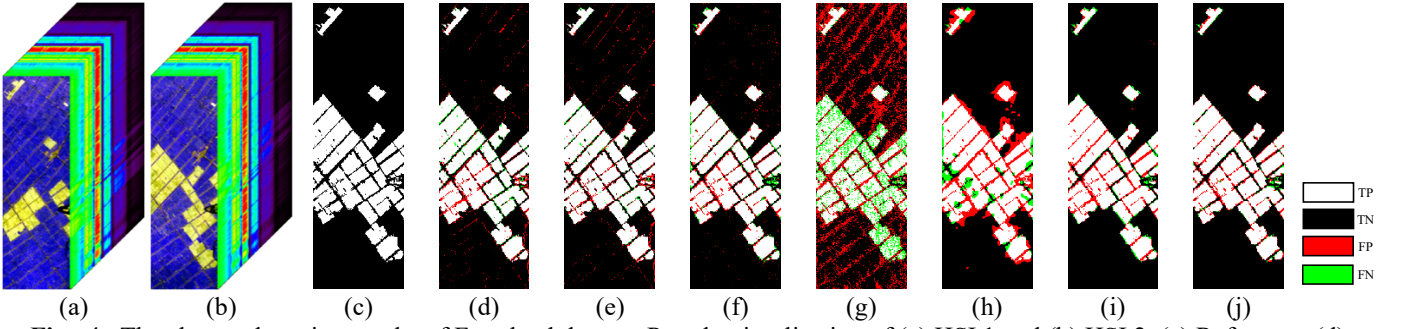

**Fig. 4.** The change detection results of Farmland dataset. Pseudo visualization of (a) HSI 1 and (b) HSI 2. (c) Reference, (d) CVA, (e) ISFA, (f) 2DCNN, (g) 3DCNN, (h) BIT, (i) CSA-Net, (j) EMS-Net.

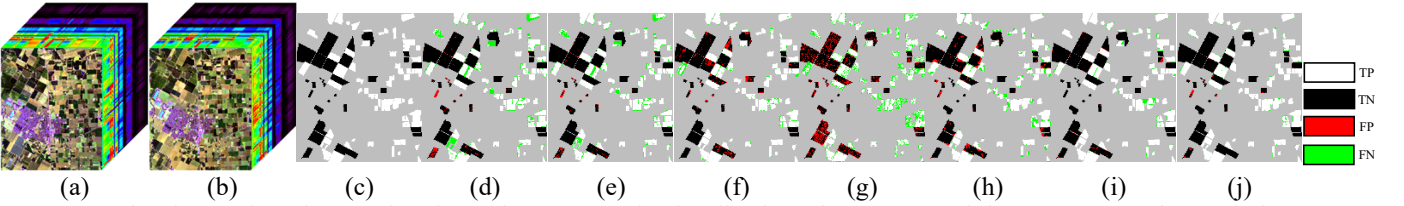

**Fig. 5.** The change detection results of Bay dataset. Pseudo visualization of (a) HSI 1 and (b) HSI 2. (c) Reference, (d) CVA, (e) ISFA, (f) 2DCNN, (g) 3DCNN, (h) BIT, (i) CSA-Net, (j) EMS-Net.

where $Q$, $K$, and $V$ denote Query, Key, and Value, respectively, and are acquired by linear projections of $\mathcal{F}_d$.

The training loss is specially designed to enhance the compactness of learned features. In details, under limited training samples, the classic cross entropy loss can only separate the changed from unchanged while neglect the compactness of two clusters. The designed supervised contrastive loss $\mathcal{L}_{SupCon}$ is defined as follows:

$$\mathcal{L}_{CroEtp} = \sum_{i \in I_U} \frac{-1}{|P(i)|} \sum_{p \in P(i)} \log \frac{exp(z_i \cdot z_p / \tau)}{\sum_{a \in A(i)} (z_i \cdot z_a / \tau)} \quad (4)$$

where $I_U$:{Unchange Set}, $P(i)$:$\{z_j, y_j = 0, z_u | Positive\ Set\ of\ z_i\}$, $z_u = \frac{1}{N_u}\sum_j z_j$, $A(i)\{z_j, y_j = 0, 1\ |All\ Set\ of\ z_i\}$, respectively.

For $\mathcal{L}_{SupCon}$, only the feature embeddings of the unchanged ones are opted for computation, where all of them and the center of the unchanged cluster are forced to be as close as possible to each other. There are various clusters for the changed ones, so the feature vectors of changed ones are excluded.

The cross-entropy loss $\mathcal{L}_{CroEtp}$ is defined as equation (5). And the final loss is calculated according to equation (6).

$$\mathcal{L}_{CroEtp} = \frac{1}{N}\sum_j -[y_j \log \hat{y}_j + (1 - y_j)\log(1 - \hat{y}_j)] \quad (5)$$

$$\mathcal{L} = \mathcal{L}_{CroEtp} + \mathcal{L}_{SupCon} \quad (6)$$

## 3. EXPERIMENTS

### 3.1 Data Description

Two real hyperspectral datasets are selected to test the effectiveness of proposed EMS-Net.

1) **Farmland dataset**: shot on May 3, 2006 and April 23, 2007 by Hyperion, in the city of Yuncheng, Jiangsu, China over farmland area. The image size is 450 × 140, with 155 spectral bands.

2) **Bay dataset**: taken on 2013 and 2015, individually, with the AVIRIS sensor surrounding the city of Patterson (California). Bay dataset is with spatial size as 600 × 500 pixels and 224 spectral bands.

The experiment is implemented on pytorch. The training samples are chose from the reference, including 100 unchanged and 100 changed samples for all datasets. The setting of the hyperparameter is that $C' = C // 8$, greatly decreasing the number of redundant feature maps and fusing change information efficiently. The training epoch is set as 200, and 0.0005 as the initial learning rate with cosine decay. Another six comparative methods are opted for comparison, namely, Change vector analysis (CVA) [16], Iterative Slow Feature Analysis (ISFA) [17], 2DCNN, 3DCNN, BIT [18], and CSA-Net [15], where BIT and CSA-Net are two newest transformer-based change detection algorithms. The evaluation index involves Overall Accuracy (OA), Kappa, F1 score, Precision and Recall rate.

### 3.2 Results and Analysis

The change detection results of Farmland dataset are presented in Fig. 4. The false alarms are in red while the omissions are in green for convenience. Compared with other methods, EMS-Net is able to detect most of change areas with little false detection and noise, especially with only a small number of samples, demonstrating the validity of extracted efficient multi-temporal change features. Fig. 5 represents the binary change maps of Bay area dataset. Little green omissions can be found in the result of proposed EMS-Net. And the unchanged and changed are mainly detected accurately. The designed supervised contrastive loss plays a role of clustering the unchanged, increasing the gap between the unchanged and changed. TABLE I gives quantitative assessment of the change detection results on Farmland dataset. The highest is bolded in red while the second one is shown in blue italics. The EMS-Net obtains comparative OA, Kappa, F1, and precision as the top one acquired by ISFA. As shown in TABLE II, EMS-Net outperforms the other comparative algorithms, yielding the largest OA, Kappa, F1, and precision, indicating the effectiveness of proposed EMS-Net.

TABLE I. THE QUANTITATIVE ASSESSMENT ON FARMLAND

| Method | OA | Kappa | F1 | Recall | Precision |
|---|---|---|---|---|---|
| CVA | 0.9548 | 0.8926 | 0.9249 | **0.8923** | 0.9601 |
| ISFA | **0.9575** | **0.8996** | **0.9301** | *0.8889* | **0.9752** |
| 2DCNN | 0.9508 | 0.8835 | 0.9187 | 0.8831 | 0.9572 |
| 3DCNN | 0.7240 | 0.4079 | 0.6109 | 0.5169 | 0.7466 |
| BIT | 0.8683 | 0.7044 | 0.8006 | 0.7136 | 0.9118 |
| CSA-Net | 0.9422 | 0.8641 | 0.9056 | 0.8609 | 0.9552 |
| EMS-Net | *0.9560* | *0.8961* | *0.9276* | 0.8865 | *0.9728* |

TABLE II. THE QUANTITATIVE EVALUATION ON BAY

| Method | OA | Kappa | F1 | Recall | Precision |
|---|---|---|---|---|---|
| CVA | 0.8723 | 0.7462 | 0.8708 | 0.9474 | 0.8057 |
| ISFA | 0.8917 | 0.7848 | 0.8905 | **0.9695** | 0.8234 |
| 2DCNN | 0.9102 | 0.8194 | 0.9166 | 0.9229 | 0.9103 |
| 3DCNN | 0.7009 | 0.4034 | 0.7053 | 0.6698 | 0.7448 |
| BIT | 0.8731 | 0.7456 | 0.8791 | 0.8637 | 0.8951 |
| CSA-Net | *0.9568* | *0.9135* | *0.9588* | 0.9395 | *0.9789* |
| EMS-Net | **0.9737** | **0.9472** | **0.9752** | *0.9679* | **0.9827** |

## 4. CONCLUSION

To conclude, a novel Efficient Multi-temporal Self-attention network is put forward to dig the relationship between various change information and reduce the redundancy, largely increase the efficiency of extracted change features under limited training samples. The experiments have been conducted on two real-world hyperspectral datasets. The results and analyses confirm the effectiveness of the proposed EMS-Net compared with the state-of-the-art approaches.

## ACKNOWLEDGMENT

This work was supported in part by the National Key Research and Development Program of China under grant 2022YFB3903302, and in part by the National Natural Science Foundation of China under grant T2122014 and 61971317.